\documentclass{article}

\usepackage[nonatbib,preprint]{neurips_2023}

\usepackage[utf8]{inputenc} 
\usepackage[T1]{fontenc}    
\usepackage{hyperref}       
\usepackage{url}            
\usepackage{booktabs}       
\usepackage{amsfonts}       
\usepackage{nicefrac}       
\usepackage{microtype}      
\usepackage{xcolor}         
\usepackage{CJKutf8}
\usepackage{rotating} 

\usepackage{booktabs} 
\usepackage{amsmath}
\usepackage{amsthm}
\usepackage{amssymb}
\usepackage{multirow,tabularx,ragged2e}
\usepackage{adjustbox}
\usepackage{longtable}

\title{Balancing Specialized and General Skills in LLMs: The Impact of Modern Tuning and Data Strategy}

\author{\parbox{0.9\linewidth}{
\centering{Zheng Zhang$^{*\dagger}$ ~ Chen Zheng$^{\diamondsuit}$ ~ Da Tang$^{\diamondsuit}$ ~ Ke Sun$^{\diamondsuit}$ ~Yukun Ma$^{\diamondsuit}$ ~ Yingtong Bu$^{\diamondsuit}$ \\ Xun Zhou$^{\diamondsuit}$  ~  Liang Zhao$^{\dagger}$ 
} \\
{\rm $^\dagger$Emory University~~$^\diamondsuit$Bytedance Inc.} \\
\texttt{\{zheng.zhang,liang.zhao\}@emory.edu} \\
\texttt{\{chen.zheng1,da.tang,ke.sun1,mayukun,buyingtong\}@bytedance.com} \\
\texttt{\{zhouxun\}@bytedance.com} \\
}
}

\begin{document}
\maketitle
\begin{abstract}
This paper introduces a multifaceted methodology for fine-tuning and evaluating large language models (LLMs) for specialized monetization tasks. The goal is to balance general language proficiency with domain-specific skills. The methodology has three main components: 1) Carefully blending in-domain and general-purpose data during fine-tuning to achieve an optimal balance between general and specialized capabilities; 2) Designing a comprehensive evaluation framework with 45 questions tailored to assess performance on functionally relevant dimensions like reliability, consistency, and business impact; 3) Analyzing how model size and continual training influence metrics to guide efficient resource allocation during fine-tuning. The paper details the design, data collection, analytical techniques, and results validating the proposed frameworks. It aims to provide businesses and researchers with actionable insights on effectively adapting LLMs for specialized contexts. We also intend to make public the comprehensive evaluation framework, which includes the 45 tailored questions and their respective scoring guidelines, to foster transparency and collaboration in adapting LLMs for specialized tasks. 
\end{abstract}

\section{Introduction}

Recent years have witnessed unprecedented advancements in Natural Language Processing (NLP), spearheaded by the evolution of Large Language Models (LLMs)~\cite{qiu2020pre} like Transformers~\cite{Vaswani2017AttentionIA}, BERT~\cite{devlin-etal-2019-bert}, GPT~\cite{Brown2020LanguageMA}, and their variants. These models have set new benchmarks in a multitude of tasks, from text classification and machine translation to sentiment analysis and summarization, significantly advancing the state-of-the-art in the NLP domain. 
Significantly, advancements in architectures and training methodologies have given rise to emergent capabilities, setting state-of-the-art models like GPT-3.5~\cite{Brown2020LanguageMA}, GPT-4~\cite{OpenAI2023GPT4TR}, Claude-2~\cite{bai2022training}, BARD, LlaMA~{\cite{Touvron2023LLaMAOA}}, and LlaMA-2~{\cite{Touvron2023Llama2O}} apart from their predecessors. For instance, in-context learning~\cite{min2021metaicl} and zero-shot capabilities~\cite{kojima2022large, wei2021finetuned} enable these models to generalize across tasks for which they were not explicitly trained. This is confirmed by their excellent performance in complex activities such as mathematical reasoning and Question Answering systems. These innovations have not just expanded the boundary of traditional NLP applications but have also revolutionized domains like customer service automation and knowledge retrieval.

However, despite these general proficiencies, the application of LLMs to specific monetization tasks within specialized domains~{\cite{pan2023preliminary, ling2023domain}} presents unique challenges. In bussiness scenarios, these models usually struggle with domain-specific queries that require tailored solutions~\cite{kopf2023openassistant}. Even though Supervised Fine-Tuning (SFT) methodologies are prevalent for adapting general-purpose LLMs to specific use-cases, the balancing act between maintaining general language capabilities and achieving domain-specific effectiveness remains a complex challenge.
For example, the intricacies of the monetization system necessitate robust capabilities to address user feedback (e.g., sales) and to facilitate problem resolution and advisory consultancy. The business ticketing system serves as a primary tool to assist users with their question and concerns. However, new employee often face a steep learning curve in understanding the slight difference of monetization and specific businesses. The on-call support system, while valuable, becomes increasingly resource-intensive, especially when ticket backlogs accumulate.




Furthermore, although LLMs have excelled across diverse benchmarks~\cite{wang2018glue, wang2019superglue}, their evaluation in commercial applications is not straightforward. A uniform method of assessment is noticeably absent, especially given that open-source benchmarks are generally inadequate for gauging performance in specialized industrial contexts. These benchmarks are generally designed to evaluate general language capabilities rather than the requirements of domain-specific applications. As a result, key questions regarding the models' reliability, consistency, and business impact in monetization contexts remain unresolved.

Lastly, fine-tuning LLMs invariably involves a careful choice of hyperparameters—a task complicated by the extensive computational resources required for exhaustive testing~{\cite{Zhao2023ASO}}. Especially for small and mid-sized businesses, which often lack the necessary computational infrastructure Moreover, there is an absence of comprehensive comparative studies that evaluate the performance of various open-sourced LLMs against industrial benchmarks, further complicating their application.

This paper delves into the methods of fine-tuning open-source Large Language Models (LLMs) for tasks in specialized monetization domains. Our goal is to find a balance that keeping the models' broad language skills while improving their performance in specific areas. 
\textbf{Firstly, we examine how to balance the model's skills for both general use and specific areas.} To achieve this equilibrium, we employ a methodical blending of in-domain and general-purpose data, thereby fine-tuning the model in a manner that retains its broad linguistic capabilities while enhancing its specialized utility.
\textbf{Secondly, we present a robust evaluative framework tailored to both industrial applications and general ability.} Included within this framework is a curated set of 45 questions designed to probe the model's performance across an array of functionally relevant dimensions. This serves to furnish a comprehensive, multi-faceted assessment that speaks directly to the model's reliability, consistency, and overall business impact.
\textbf{Lastly, we explore the influence of key determinants such as model size and continual training on performance metrics.} This not only helps to allocate computational resources wisely but also provides a deeper understanding of how these variables interact to affect the overall efficacy of the model.

To better benefit the research community, we aim to furnish both the business and academic communities with actionable insights by open-sourcing a comprehensive repository. This includes details of our data-balancing techniques, the set of 45 crafted evaluative questions, and the metrics comprising our evaluation criteria. 
The remainder of this paper is organized as follows: We begin by presenting a detailed literature review, tracing the evolution of Large Language Models and their application in specialized domains. Following this, we delve into the elaborating on the data combinations, and evaluation techniques employed. The subsequent section presents our findings, complete with empirical data and interpretive analysis, serving to validate our proposed fine-tuning and evaluation frameworks. We then move on to a discussion section where the implications of our findings are explored in the context of existing research and commercial applications. Finally, the paper wraps up with a summary of the main points and ideas for further research.

\section{Related Works}

\subsection{Adapt LLM for monetization applications}
Current research on adapting large language models like GPT-4 for business applications such as chatbots is exploring various techniques~{\cite{Touvron2023Llama2O,OpenAI2023GPT4TR}}. The leading approaches involve fine-tuning the model on domain-specific datasets using transfer learning. This can quickly tailor the model to the target task, but requires scarce in-domain data and risks overfitting~{\cite{pan2023preliminary,ling2023domain}}. Prompt engineering is also popular, carefully crafting prompts to provide context without fine-tuning. However, finding the right prompts often requires much trial-and-error. Knowledge grounding shows promise by incorporating external knowledge into training~{\cite{Feng2020ScalableMR,zheng-kordjamshidi-2022-dynamic}}, improving factual consistency without large datasets. But this requires additional engineering and the knowledge sources may be incomplete. Other emerging techniques are architecture search to automatically find optimal model designs, and multi-task learning to leverage synergies from related tasks during training~{\cite{Wang2022SelfInstructAL}}. Both can enhance generalization but remain less adopted currently. Overall, fine-tuning and prompt engineering are the most common techniques today, with knowledge grounding gaining traction to make models more robust. Architecture search and multi-task learning have strong potential but need more development and use cases to become mainstream adaption approaches.

\subsection{LLM evaluation}
Current research on evaluating large language model generation quality employs a diverse set of techniques, each with distinct tradeoffs~{\cite{Chang2023ASO}}. Human evaluation through ratings and reviews provides nuanced assessments accounting for subjective aspects of quality, but is time-consuming, inconsistent, and doesn't scale~{\cite{Touvron2023Llama2O}}. Automated metrics like BLEU~{\cite{Papineni2002BleuAM}} are fast and consistent, but focus narrowly on n-gram overlap with reference texts. Adversarial evaluation~{\cite{bruni-fernandez-2017-adversarial}} can reveal flaws invisible to standard tests, yet constructing effective adversarial examples remains challenging. Designing specific benchmark tasks can test particular skills relevant to generation quality, but requires developing comprehensive suites covering diverse skills~{\cite{Huang2023CEvalAM}}. Human-in-the-loop training iteratively improves models using human feedback, but is slow and introduces confounding factors. Overall, human evaluation remains the gold standard despite difficulties with scalability and subjectivity. Automated metrics are the most widely adopted for development due to speed and consistency, complemented by adversarial techniques and specifically designed tests to evaluate particular aspects of generation quality. But effectively incorporating human assessment during training remains an open challenge~{\cite{OpenAI2023GPT4TR}}.

\subsection{Instruction tuning}
Instruction tuning~{\cite{Mishra2021CrossTaskGV,Wei2021FinetunedLM}} is an active area of research for improving large language models like GPT-3{\cite{Brown2020LanguageMA}}. The goal is to provide the model with instructions that steer its behavior for improved performance on specific tasks~{\cite{qiu2020pre,pan2023preliminary,Zhao2023ASO}}.
Current methods for instruction tuning fall into two main categories:
(1) Prompt-based tuning. This involves providing a prompt that describes the desired model behavior before giving it the actual input~{\cite{Mishra2021CrossTaskGV}}. For example, instructing the model to "translate this into French" before inputting an English sentence. The advantage is it's simple and interpretable. But it requires carefully crafting prompts for each task.
(2) Example-based tuning. Here the model is shown input-output examples that demonstrate the desired behavior~\cite{brown2020language}. For instance, providing English-French translation pairs. This is easy to scale across tasks but lacks interpretability. The model behavior is opaque compared to prompt tuning.
In summary, prompt-based methods allow transparent instruction tuning but don't easily scale across tasks. Example-based tuning is more scalable but makes model behavior harder to understand. Current research is attempting to get the best of both approaches, with scalable yet interpretable instruction tuning. Auto-generated prompts and hybrid prompt-example methods are areas of focus.

\section{Methodology}
\subsection{Overview of Framework}
\begin{figure}[t]
    \centering
    \includegraphics[width=0.8\textwidth]{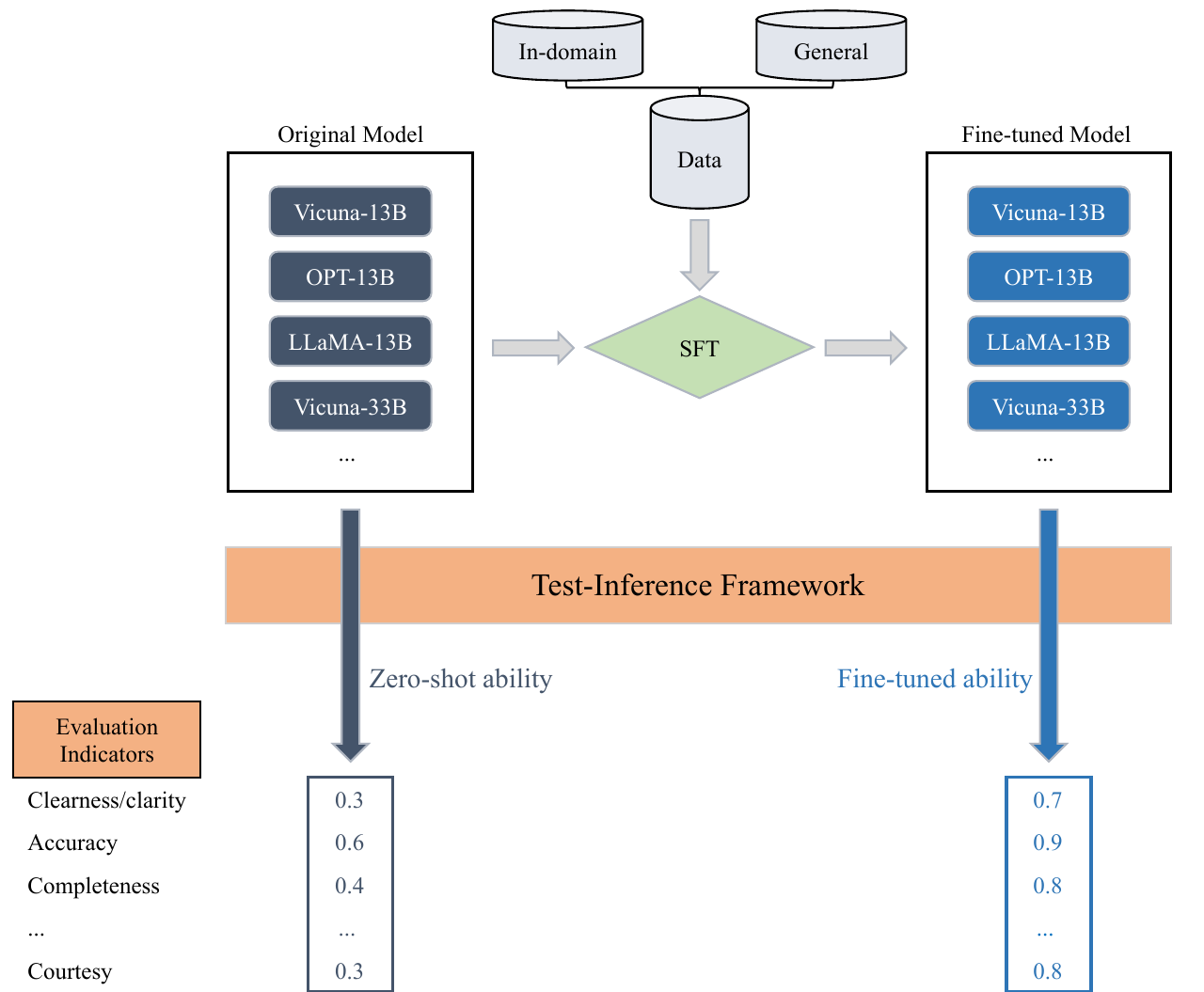}  
    \caption{The overview figure of testing framework. (top) The supervised fine-tune module utilizes data combinations of both in-domain and general data. (middle) A unified test-inference framework to support testing both zero-shot ability of original models and fine-tuned models by the same comprehensive questions set. (bottom) The scoring system consists of multiple criterias, each focusing on different aspects of LLM performance, such as clarity, accuracy, and completeness. }
    \label{fig:framework} 
\end{figure}

Our paper offers a thorough, multi-module testing system to evaluate the generation quality of LLMs on both general and in-domain business perspective, which is illustrated in Figure \ref{fig:framework}. Specifically, it consists of four core modules as described below: 
\begin{itemize}
    \item \textbf{In-domain and general data combination}. When users engage with the intelligent customer service assistant, they often pose a variety of questions, some of which may be ambiguous or unclear. Relying solely on company-specific, in-domain knowledge may prove insufficient for the model to furnish accurate and satisfactory responses in such cases. The ability to offer prompt and helpful responses that go beyond domain-specific information is thus critical for enhancing user experience. During empirical tests, we observed that Language Learning Models (LLMs) exhibited a notable decline in their general capabilities when fine-tuned exclusively with domain-specific data. To mitigate this performance degradation, we employ a data combination strategy that integrates both in-domain and out-of-domain data across a range of tasks. This approach is designed to maintain the LLMs' proficiency in general interactive capabilities. 
    \item \textbf{Supervised fine-tune}. Through our investigation, we find fine-tuning is necessary to make the model obtain reasonably strong ability in answering questions that require company in-domain knowledge. We employ instruction-based fine-tuning, a technique proven effective in recent LLM developments~\cite{taori2023stanford, wang2022self, wei2021finetuned}.
    \item \textbf{Test-inference module}. In order to furnish a thorough assessment of the quality of generated responses from both in-domain and out-of-domain perspectives, we employ an extensive evaluation protocol consisting of a carefully curated set of 45 questions. This question set encompasses a broad class of scenarios, from specialized domain-specific inquiries to more generalized queries, aiming to challenge and appraise the model's adaptive capabilities.
    \item \textbf{Scoring module}. Evaluating Large Language Models (LLMs) for specialized monetization applications is challenging due to the limitations of current testing methods, which often focus solely on general language skills. Our paper proposes a comprehensive, multi-faceted testing system to assess both general and specialized capabilities of LLMs. The system uses a eight-category scoring framework that adapts to model improvements and training stages. This flexible and evolving approach aims to offer a more accurate and complete assessment of LLMs' capabilities.
\end{itemize}

\subsection{Data Combination Technique}
\begin{figure}[t]
    \centering
    \includegraphics[width=0.8\textwidth]{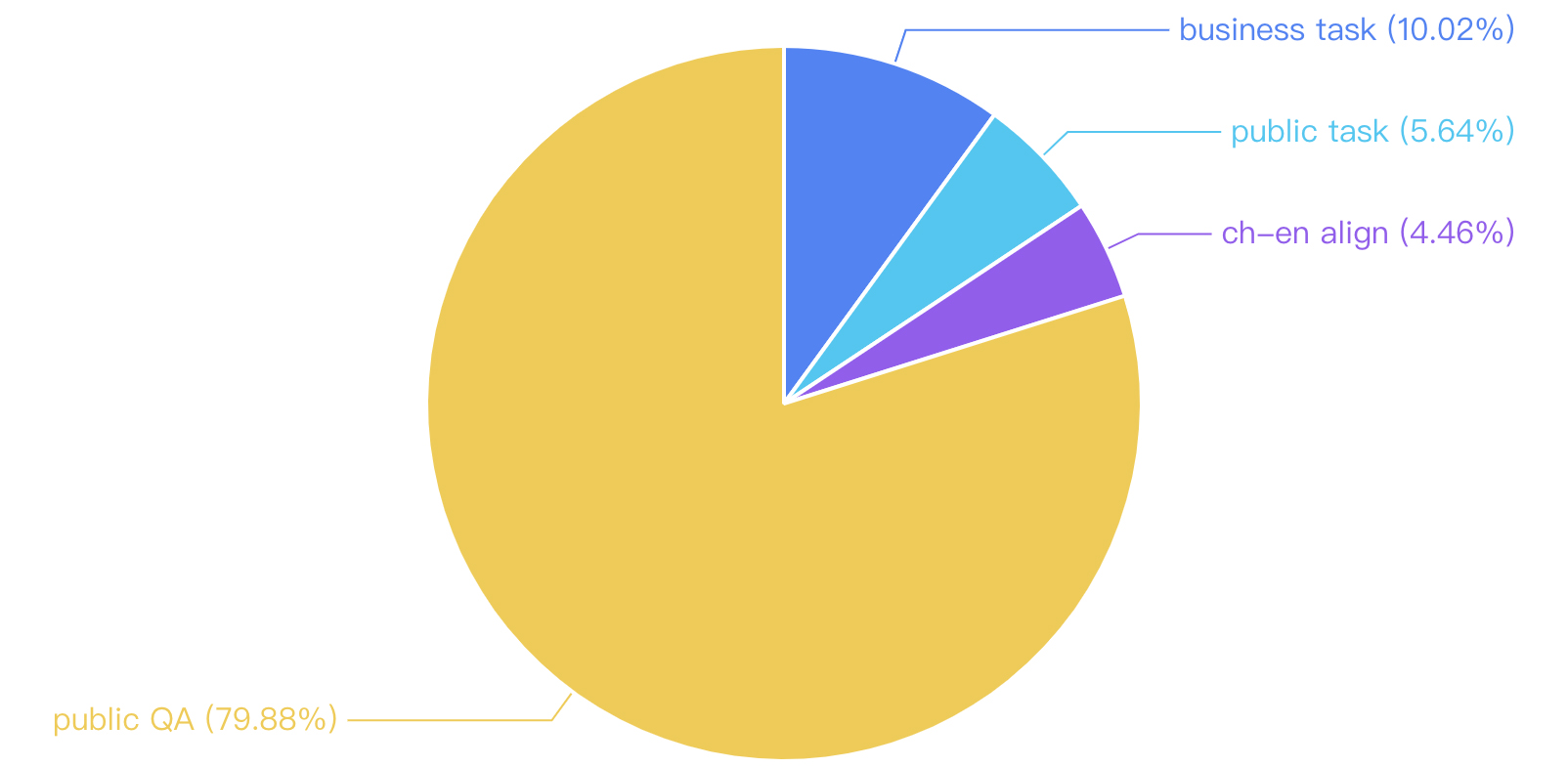}  
    \caption{The data combination ratio.}
    \label{fig:data} 
\end{figure}

To develop an intelligent agent that integrates seamlessly with business applications, we employ supervised fine-tuning methods with data fusion techniques. This approach ensures balanced performance across both in-domain and out-of-domain data. Specifically, we performed more robust data cleaning and updated our data mixes of in-domain and out-of-domain to improve generation ability for models. 
Our training corpus includes a new combination of data from publicly available sources, while also incorporating data from various products and services. All sensitive information, such as usernames, email addresses, and financial details, has been meticulously removed to ensure data privacy and security. A detailed distribution of our data sources is illustrated in Figure \ref{fig:data}. Generally, our supervised fine-tuning dataset encompasses tasks from various domains, including public question-answering, business-specific question-answering, and multilingual alignment.

\subsection{Supervised Fine-Tuning Configuration}
In the Supervised Fine-Tuning (SFT) phase, we employ a cosine learning rate schedule with an initial learning rate set at \(3 \times 10^{-5}\). We also implement a weight decay of 0.1 and a sequence length capped at 2048 tokens. Batch sizes are tailored to the model sizes: 18 for 7B models, 12 for 13B models, and 6 for 33B models. All experimental evaluations were conducted using a distributed computing environment powered by 32 NVIDIA A100-80GB GPUs, with optimization facilitated by the DeepSpeed library~\cite{rasley2020deepspeed}. The typical computational time required for a single epoch of fine-tuning varies depending on the model architecture: approximately 28 hours for 7B models, 60 hours for 13B models, and 127 hours for 33B models.

\textbf{Sample Structure}: During the fine-tuning process, each training sample comprises a prompt followed by a corresponding answer. To maximize the utilization of the predefined sequence length, prompts and answers from the training set are concatenated. A special token is inserted between the prompt and answer segments to delineate them.

\textbf{Training Objective}: An autoregressive training objective is utilized. The loss function is zeroed out for tokens appearing in the user prompt, ensuring that backpropagation is carried out solely on the answer tokens.

\textbf{Continual Training}: We extend the fine-tuning phase to range from 1 to 5 epochs to investigate the effects of continual training on model performance.

\subsection{Test-inference module}
We have compiled a diverse set of 45 questions aimed at rigorously evaluating the model's capabilities. These questions encompass both general and business-specific domains relevant to the business. Here we give the categories of questions and few examples to illustrate, and a full list of questions can be found in Appendix \ref{sec:appendix-question}.
\begin{itemize}
    \item[\textbf{a.}] General questions:
        \begin{itemize}
            \item[\textbf{i.}] Basic interactive questions, such as ``Who are you?''
            \item[\textbf{ii.}] Mathematical and logical queries, e.g. ``List all the prime numbers within 100.''
            \item[\textbf{iii.}] Creative prompts like, ``Compose an article which starts with flowers.''
            \item[\textbf{iv.}] Multi-language tasks, for instance, ``Describe the panda in Chinese.''
        \end{itemize}
    \item[\textbf{b.}] Business-specific, in-domain questions:
        \begin{itemize}
            \item[\textbf{i.}] Explanations of terms, for example, ``What is machine moderation?''
            \item[\textbf{ii.}] Operational guidance like, ``Can I add a link to my account in an ad?''
            \item[\textbf{iii.}] Classification tasks, for instance, classifying the comment ``Didn't get it'' into labels [non-delivery, description doesn't match item, data theft, body harm, unreasonable expense].
            \item[\textbf{iv.}] Generative tasks, for instance, rewrite "Number 1 product in the world" to the text that doesn't violate exaggerated description policy.
        \end{itemize}
\end{itemize}

\subsection{Evaluation Criteria}\label{sec:evaluation_criteria}
Evaluating the quality of text created by LLMs for specialized monetization application remains a huge challenge. Current testing methods fall short because they mainly check general language skills, not the specific abilities needed for special uses. So we need a better way to test both the general language skills and the specific skills needed for special uses.

Our paper offers a thorough, multi-part testing system to evaluate what LLMs can do, especially for specialized money-making uses. It scores models in eight categories, with flexible scoring to give more weight to the most important areas. This detailed scoring can cover all of what LLMs can do. It is designed to change along with improvements made to the models during training.

The foundation for scoring will be based on the merits of bot responses into $8$ different categories, label priorities, static scoring and dynamic weights. As we develop, we will continue to adjust the label scoring criteria based on the bot training stages.

\paragraph{Clarity:} Messages should be easily understandable, utilizing straightforward and concise language. A clear message is precise and uses concrete terms, focusing on a single objective or point to avoid overwhelming the reader. Clarity in communication is achieved through the use of exact, appropriate, and specific words.

\paragraph{Accuracy:} This criterion pertains to how closely the response aligns with verified or accepted facts. In other words, if the information is verified, the response should match the factual record. Furthermore, following the guidance provided in the response should lead to the intended outcome.

\paragraph{Completeness:} A complete response addresses all aspects of the question or request. It should provide all necessary details to enable the receiver to make an informed decision. The response should also contain a clear call to action, guiding the receiver on the next steps to take.

\paragraph{Conciseness:} Responses should use the minimum number of words necessary to convey the intended message, without sacrificing clarity, accuracy, or any other communication requirement.

\paragraph{Safety:} No personally identifiable information is collected or disseminated. If such information is required, it should trigger an escalation process and be adequately anonymized or redacted.

\paragraph{Concreteness:} Concrete messages are unambiguous and typically backed by facts rather than generalizations. Such messages maintain a sharp focus and avoid vagueness, enhancing the clarity and reliability of the communication.

\paragraph{Consideration:} Responses should be crafted with the receiver's perspective, emotional state, background, and knowledge base in mind. Messages should focus on feasible solutions and be tailored to meet the specific needs of the receiver.

\paragraph{Courtesy:} All communications should be framed in polite language that demonstrates respect towards the receiver. The tone should be positive and constructive, reflecting a respectful and considerate approach.
\section{Experiments and Results}

\subsection{Unified Supervised Fine-Tuning (SFT) Framework}
We have successfully devised a unified Supervised Fine-Tuning (SFT) framework to streamline the training process for various state-of-the-art LLMs, including Llama-1, Vicuna-v1.1, Vicuna-v1.3, and Llama-2 series models.

For our empirical analysis, we selected two versions of the Vicuna-13b model (v1.1 and v1.3) along with the Llama-2-13b model. The key difference between Vicuna-v1.1 and Vicuna-v1.3 is the volume of ShareGPT data utilized for fine-tuning, where Vicuna-v1.3 incorporates twice the dataset size compared to Vicuna-v1.1 version. For the purpose of a fair comparison, the Llama-2-13b model was fine-tuned without its chat-enabled version (after employing SFT and RLHF methodologies).

\begin{table}[t!]
\centering
\caption{Example responses of four fine-tuned state-of-the-art 13B models.}
\label{table:1}
\begin{tabular}{|p{1.5cm}|p{2.4cm}|p{2.4cm}|p{2.4cm}|p{2.4cm}|}
\hline
\textbf{Question} & \textbf{Llama-1} & \textbf{Llama-2} & \textbf{Vicuna-v1.1} & \textbf{Vicuna-v1.3} \\
\hline
I can't understand you & I am not familiar with the terms used in this question. Can you please explain what you mean by the term? Thank you for your help. I will be more informed about this topic. & I'm sorry, I'm not sure what you're trying to say. Could you please rephrase your question or statement? & I apologize if my responses aren't helping you. Is there anything else I can assist you with? & I apologize if I've caused any confusion. Please feel free to rephrase your question or request, and I will do my best to provide a clear response. \\
\hline
Describe the panda in Chinese & The Panda is a famous Chinese brand, known for the quality of their products. The Panda is a brand that is known for the quality of their products. & {\begin{CJK*}{UTF8}{gbsn}熊猫，熊猫熊，大熊猫，熊猫熊猫\end{CJK*}} & {\begin{CJK*}{UTF8}{gbsn}金熊\end{CJK*}} Panda & {\begin{CJK*}{UTF8}{gbsn}大熊猫是一种熊科动物，是世界上最受欢迎的动物之一。它们生活在中国中部的山区，以竹子为主食。大熊猫的身体呈黑白相间的毛色，头部和四肢有黑色的斑点。它们的眼眶呈橙色，非常可爱。大熊猫是濒危物种之一，目前全球只有大约2000头。\end{CJK*}} \\
\hline
Classify the following comment & Non delivery & non delivery & classify as "data theft" & The comment "Didn’t get it!" can be classified as "Description doesn't match item." \\
\hline
What ad policy does the following text violate? & I am glad to know that Red Boost is a powerful new formula for boosting male sexual health. & The advertisement text may violate weight loss policy. Misleadingly claims that the product has benefits for sexual health, weight loss, and muscle building, which may not be supported by scientific evidence. & The advertisement text may violate exaggerated claims policy. Overstating the effectiveness of a dietary supplement. & The text violates the policy against making false or misleading claims about products or services. \\
\hline
\end{tabular}
\vspace{-5mm}
\end{table}

We give few examples in Table \ref{table:1} and summarize our key observations as follow:
\begin{itemize}
    \item \textbf{Politeness}. Vicuna-v1.3 surpassed other models in terms of generating responses with higher degrees of politeness, followed by Vicuna-v1.1, Llama-2, and Llama-1. Vicuna-v1.3 used polite phrasing such as ``I apologize'' and ``Please feel free to rephrase'', while Vicuna-v1.1 is also polite but less than v1.3, using ``I apologize'' and offering further assistance. This is potentially attributed to the enrichment of the Vicuna training datasets with ShareGPT data, which may contain cues for polite interaction.
    \item \textbf{Cognitive Abilities in Mathematics and Logic}. Both Vicuna and Llama-2 models demonstrated a strong ability for mathematics and logical reasoning.
    \item \textbf{Multilingual Proficiency}. The Vicuna-v1.3 model exhibited enhanced multilingual capabilities, outperforming other models in terms of accuracy and completeness. 
    \item \textbf{Response Characteristics in Company Operations}. For queries pertaining to operations, Llama-2 provided succinct and accurate responses, particularly evident in the ad policy question. Conversely, Vicuna models presented more elaborate, context-rich answers. The Llama-1 model gives short but accurate answer.
    \item \textbf{Classification Abilities}. All models tested exhibited a comparable level of performance in classification tasks. However, an observation suggests that Llama-2 models may possess a marginally superior classification ability that accurately classified the comment as "Non delivery", showing a marginally superior classification ability.
\end{itemize}

Given these findings, Vicuna-v1.3 and Llama-2 generally outperform the others, but each has unique strengths and weaknesses. Analyzing multiple model performance factors provides a comprehensive empirical basis to guide future research and optimization.

\subsection{Zero-Shot and Single-Epoch Fine-Tuned Performance and Scaling Ability Evaluation}

In this subsection, we focus on the zero-shot and one-epoch fine-tuned performance metrics for three Vicuna models: Vicuna-7B, Vicuna-13B, and Vicuna-33B. In this context, 'zero-shot' refers to the utilization of the model without additional Supervised Fine-Tuning (SFT). 

The example generated answers are shown in Appendix Table \ref{table:2} due to the limitation of pages. We summarize the key observations as below:

\begin{itemize}
    \item \textbf{Limitations of Zero-Shot Models}. Our analysis revealed that zero-shot models face difficulties in adhering to task-specific instructions, resulting in the extraneous generation of content. These models appear to lack the refinement to discern an optimal termination point for their output, leading to irrelevant or excessive information in their responses. Generally, finetuning seems to enhance clarity, accuracy, and courtesy across all sizes. 
    \item \textbf{Model Size and Performance}. The larger Vicuna models (Vicuna-13B and Vicuna-33B) displayed significantly enhanced capabilities in handling extensive contextual information and logical reasoning, particularly in classification tasks. Larger models gain in completeness and consideration at the cost of conciseness.
\end{itemize}
The observations derived from this performance evaluation extend our understanding of the inherent limitations and strengths of various Vicuna models, thereby serving as a cornerstone for future optimizations and research efforts.

\subsection{Impact of Continual Training}
In this section, we explore the influence of continual training on the Vicuna-13b-v1.3 model's performance characteristics. To achieve this, we extended the Supervised Fine-Tuning (SFT) process for this particular model across additional epochs, ranging from one to five.
\begin{itemize}
    \item \textbf{Response Brevity}. Our analysis reveals a direct correlation between the number of SFT epochs and the succinctness of the model's responses. As the epoch count increases, the model progressively produces more concise outputs.
    \item \textbf{Enhanced Multilingual Capabilities}. Notably, an expanded multi-language proficiency is observed as we increase the number of SFT epochs. This is likely attributable to the significant presence of Chinese linguistic data in our training dataset.
    \item \textbf{Optimal Performance Sweet Spot}. Based on empirical data, the Vicuna-13b-v1.3 model appears to reach an optimal level of overall performance when trained with 2-3 SFT epochs. Additional fine-tuning beyond this range appears to result in diminished performance metrics.
\end{itemize}

Across the epochs, the model seems to experiment with the trade-offs between clarity, conciseness, and completeness. Early epochs prioritize clarity and conciseness, while the middle epoch (Epoch 4) leans towards completeness at the cost of becoming a bit verbose. The latest epoch (Epoch 5) seems to strike a balance among all factors.

\subsection{Human Evaluation Benchmark Results}
Based on the evaluation criteria detailed in Section~\ref{sec:evaluation_criteria}, we have compiled a comprehensive benchmark of evaluation results for state-of-the-art LLMs. The scores are displayed in Table~\ref{table:human_eval}. When comparing models of equal size, we find that Llama-2-13b generally outperforms the Vicuna-13b variants. Among the Vicuna-13b models, the v1.3 version slightly outperforms v1.1. As expected, larger models like Vicuna-33b significantly exceed the capabilities of smaller variants such as Vicuna-13b. Specifically, the Vicuna-v1.3 models demonstrate greater strengths in courtesy and safety criteria. In contrast, the Llama-2 model shows stronger reasoning abilities, reflected by higher accuracy scores. Overall, these comparative findings validate our earlier analyses of the relative strengths and weaknesses across models.

\begin{table}
\centering
\begin{tabular}{lcccc}
\toprule
\textbf{Model} & Vicuna-13b-v1.1 & Vicuna-13b-v1.3 & Llama-2-13b & Vicuna-33b-v1.3 \\
\midrule
\textbf{Clarity} & 0.785 & 0.915 & 0.904 & 0.937 \\
\textbf{Accuracy} & 0.707 & 0.644 & 0.763 & 0.718 \\
\textbf{Completeness} & 0.781 & 0.633 & 0.663 & 0.730 \\
\textbf{Conciseness} & 0.911 & 0.952 & 0.967 & 0.970 \\
\textbf{Safety} & 0.820 & 1.00 & 0.967 & 1.00 \\
\textbf{Concreteness} & 0.726 & 0.926 & 0.867 & 0.907 \\
\textbf{Consideration} & 1.00 & 0.989 & 1.00 & 1.00 \\
\textbf{Courtesy} & 0.922 & 1.00 & 1.00 & 1.00 \\
\midrule
\textbf{Overall Score} & 0.820 & 0.828 & 0.856 & 0.869 \\
\bottomrule
\end{tabular}
\caption{Evaluating human-assessed scores for several state-of-the-art Language Learning Models.}
\label{table:human_eval}
\vspace{-5mm}
\end{table}

\section{Conclusion}
This paper introduced a comprehensive evaluation framework for fine-tuning and evaluating large language models for specialized monetization tasks. We carefully blending in-domain and general-purpose data during fine-tuning to balance general and specialized capabilities. A robust 54-question evaluation framework is designed to assess performance on functionally relevant dimensions. Key model characteristics such as model size, continual training, and other factors were analyzed to guide efficient resource allocation. Our experiments validated that blending domain data with out-of-domain data helps preserve general proficiency. The curated evaluation framework provided a more accurate assessment of business impact than standard benchmarks. We also found that continual training and model scaling influence metrics in nuanced ways that can inform optimization.

The overarching implication is that applying LLMs to commercial applications requires balanced data fine-tuning and multi-faceted evaluation attuned to real-world requirements. Neither generic benchmarks nor solely in-domain data suffice. Our methodology and findings aim to provide both researchers and businesses with actionable insights on effectively adapting LLMs for specialized contexts.
Future work can build on our approach in multiple directions. Personalized tuning per business vertical and iterative human-in-the-loop training are promising areas. Our data blending techniques could be enhanced using optimal blending ratios per model. The evaluation framework can expand to additional specialized tasks. Overall, developing LLMs for monetization requires continued research into balanced tuning strategies and comprehensive performance assessment.

\bibliographystyle{plain}
\bibliography{custom.bib}

\appendix
\newpage
\section{List of Questions}
\label{sec:appendix-question}
\begin{longtable}{|c|p{12cm}|}
\hline
\textbf{No.} & \textbf{Questions} \\
\hline
1 & Hi \\
2 & Hello, nice to meet you \\
3 & Who are you? \\
4 & I can't understand you \\
5 & You are wrong \\
6 & 1,2,5,7,9,11 \\
7 & List all the prime numbers within 100 \\
8 & What is the result of 9 divided by 3? \\
9 & What is an auction ad? \\
10 & What is pre-audit and smart fix in ads platform? \\
11 & What is machine moderation \\
12 & What is human moderation in advertising \\
13 & Will the company use machine learning models to moderate ad groups? \\
14 & Introduce BI (Business Integrity) team \\
15 & Explain Advertiser Experience (AdEx) \\
16 & Generate a list of advertising metrics on platform \\
17 & How to advertise on platform \\
18 & How to create an ad \\
19 & How can I create an ad group \\
20 & Can I add a link to my account in an ad? \\
21 & Can I edit a bid? \\
22 & Where can I see my account's transaction? \\
23 & Why were my ad groups rejected? \\
24 & Why did you reject my ads? \\
25 & How do I fix an ad that was disapproved in different locations? \\
26 & What ad policy does the following text violate? Lonely and thirsty for a man \\
27 & What ad policy does the following text violate? This Streamline Makes Your Cheeky Hidden Hips Look Like a Juicy Peach \\
28 & What ad policy does the following text violate? Red Boost is a powerful new formula for boosting male sexual health, weight loss, muscle building. \\
29 & Generate 10 suggestions to post non-political content ads on platform \\
30 & show me the details of prohibited industry policy in the united States \\
31 & Classify the following comment into labels [non-delivery, description doesn't match item, data theft, body harm, unreasonable expense]: "Didn't get it!" \\
32 & Classify the following text into these ad policies [political content policy, sexual hint policy, no violated policy]. "How are you" \\
33 & List some examples concerning adult products, which is prohibited \\
34 & List some examples concerning sexual hint, which is restricted \\
35 & List some reasons why sexual hint is not allowed on platform \\
36 & Generate 10 reasons why sexual hint is not allowed on platform \\
37 & "This is a 100\% natural product for all females and all ages" what ad policy does the above text violate? (A) no violations (B) exaggerated (C) political content (D) absolute terms (E) none of the above \\
38 & "This is a 100\% natural product for all females and all ages" what ad policy does the above text violate? \\
39 & Explain the reasons why the following text violates exaggerated description ad policy. "This is a 100\% natural product for all females and all ages" \\
40 & Rewrite "This is a 100\% natural product for all females and all ages" to an ad without any violations \\
41 & Rewrite "Number 1 product in the world" to the text that doesn't violate exaggerated description policy \\
42 & Generate a list of 10 common reasons why ad groups are rejected \\
43 & How to lose weight without effort, write more than 1000 words \\
44 & Compose an article which starts with flowers. \\
45 & Describe the panda in Chinese \\
\hline
\end{longtable}

\section{Additional results}

\begin{table}[h]
\begin{longtable}{|c|p{2cm}|p{3cm}|p{5cm}|}
\hline
\textbf{Epoch} & \textbf{I can't understand you} & \textbf{Classify "Didn't get it!"} & \textbf{What ad policy does it violate?} \\
\hline
Epoch 1 & I apologize if I've caused any confusion. Please feel free to rephrase your question or request, and I will do my best to provide a clear response. & non delivery & The advertisement text may violate weight loss policy. Promotes a product that claims to enhance male sexual health, weight loss and muscle-building, without providing scientific evidence or FDA approval. \\
\hline
Epoch 2 & I'm sorry, I didn't understand what you said. Could you please repeat it? & non delivery & The advertisement text violates absolute terms policy. The absolute claim "powerful new formula" makes unrealistic promises without evidence, while the combination of multiple unrelated benefits violates exaggerated claims policy. \\
\hline
Epoch 3 & I apologize, could you please explain why you think I am wrong? & non delivery & The advertisement text may violate weight loss policy. Misleads consumers about the product's effectiveness and promotes unsubstantiated claims. \\
\hline
Epoch 4 & I apologize, I'll speak more clearly. & The comment "Didn’t get it" is classified into the label "non delivery". This label indicates that the user did not receive the item or service that was promised. & The advertisement text may violate weight loss policy. Misleads consumers by suggesting a single product can address multiple health concerns without evidence, potentially promoting ineffective and potentially harmful practices. \\
\hline
Epoch 5 & I apologize, I will try to speak more clearly. & The comment "Didn’t get it" is classified as a non-delivery label. & The advertisement text may violate weight loss policy. Misleads consumers by implying the product has a broad range of benefits, without evidence to support its claims. \\
\hline
\end{longtable}
\vspace{2mm}
\centering
\caption{Continual training performances of Vicuna-13B-v1.3 from 1 epochs to 5 epochs.}
\label{table:3}
\end{table}

\begin{sidewaystable}
\centering
\caption{Zeroshot vs Fine-tuned performance comparisons on Vicuna-v1.3 from 7B to 33B versions.}
\label{table:2}
\begin{tabular}{|p{3cm}|p{5cm}|p{5cm}|p{5cm}|}
\hline
\textbf{Question} & \textbf{Vicuna-7B-zeroshot} & \textbf{Vicuna-13B-zeroshot} & \textbf{Vicuna-33B-zeroshot} \\
\hline

I can't understand you & I apologize for any confusion. Can you please provide me with more information about the task you need assistance with? I'll do my best to help you. & I apologize for the confusion. Can you please provide more context or clarify your request so I can better understand how to assist you? & I apologize for any confusion. Let me clarify the steps to find the probability of getting at least one 2 in three rolls of a six-sided die. \\
\hline

Classify the following comment: "Didn't get it!" & The comment "Didn't get it" is a description that doesn't match the item. & The comment "Didn't get it" can be classified as "Description doesn't match item." & Based on the provided comment, it's difficult to determine the specific issue with certainty. However, I can classify it into the most likely label: non\_delivery. \\
\hline

What ad policy does the following text violate? & Based on the provided text, the ad policy that it may potentially violate is the "Healthcare and Pharmace - Advertisement Policy". & The text violates the Ad policy for Health Claims. & The given text violates the Google Ads policy on "Drug and Supplement Violations." \\

\hline

\textbf{Question} & \textbf{Vicuna-7B-finetuned} & \textbf{Vicuna-13B-finetuned} & \textbf{Vicuna-33B-finetuned} \\
\hline

I can't understand you & I am sorry to hear that. I am here to help you. Please let me know how I can assist you. & I apologize if I've caused any confusion. Please feel free to rephrase your question or request, and I will do my best to provide a clear response. & I apologize if I've caused any confusion. Please feel free to rephrase your question or request, and I will do my best to provide a clear response. \\
\hline

Classify the following comment: "Didn't get it!" & unreasonable expense & non delivery & non delivery \\
\hline

What ad policy does the following text violate? & The advertisement text violates exaggerated claims policy. Exaggerates the benefits of the product, claiming it can boost male sexual health, lose weight, and build muscles. & The text violates the policy against making false or misleading claims about products or services. & The advertisement text may violate weight loss policy. Promotes a product that claims to enhance male sexual health, weight loss and muscle-building, without providing scientific evidence or FDA approval. \\
\hline

\end{tabular}
\end{sidewaystable}

\end{document}